%% file: main.tex
\pdfoutput=1

\documentclass[11pt]{article}

\usepackage{EMNLP2022}

\usepackage{times}
\usepackage{latexsym}
\usepackage{xspace}
\usepackage[a-1b]{pdfx}
\usepackage{hyperref}
\usepackage[utf8]{inputenc}
\usepackage{colorprofiles}
\hypersetup{pdfstartview=}

\usepackage[T1]{fontenc}

\usepackage[utf8]{inputenc}

\usepackage{microtype}
\usepackage[normalem]{ulem}

\usepackage{inconsolata}

\usepackage{amssymb}
\usepackage{amsmath}
\usepackage{tabularx}
\usepackage{multirow}
\usepackage{caption}
\usepackage{subcaption}

\usepackage{nicefrac}
\usepackage{graphicx}
\graphicspath{ {./graphs/} }
\usepackage{pgf}

\newcommand{\com}[1]{}
\newcommand{\resolved}[1]{}

\newcommand{\draftcomment}[3]{{\textcolor{#3}{[#1]#2}}}
\renewcommand{\draftcomment}[3]{}

\newcommand{\appref}[1]{App.~\ref{app:#1}}
\newcommand{\subsecref}[1]{Sec.~\ref{ssec:#1}}
\newcommand{\figref}[1]{Fig.~\ref{fig:#1}}
\newcommand{\tabref}[1]{Tab.~\ref{tab:#1}}

\newcommand\addtag{\refstepcounter{equation}\tag{\theequation}}

\newcommand {\R}{\mathbb{R}}

\newcommand {\methodName} {PAPA\xspace}

\newcommand{\base}[0]{original\xspace}

\title{How Much Does Attention Actually Attend? \linebreak Questioning the Importance of Attention in Pretrained Transformers}

\author{\textbf{Michael Hassid}$^\heartsuit$ \quad\textbf{Hao Peng}$^{\diamondsuit*}$ \quad
\textbf{Daniel Rotem}$^\heartsuit$ \quad
\textbf{Jungo Kasai}$^\spadesuit$ \quad
        \textbf{Ivan Montero}$^{\bigstar}$\thanks{\hspace{1mm}This work was done while Hao Peng and Ivan Montero were at the University of Washington.}\\
       \textbf{Noah A. Smith}$^{\spadesuit\diamondsuit}$ \quad        \textbf{Roy Schwartz}$^\heartsuit$
       \\
  $^\heartsuit$School of Computer Science \& Engineering, Hebrew University of Jerusalem\\
  $^\diamondsuit$Allen Institute for Artificial Intelligence\quad$^\bigstar$Apple, Inc.\\ 
  $^\spadesuit$Paul G. Allen School of Computer Science \& Engineering,
  University of Washington\\
{\tt \{michael.hassid,daniel.rotem,roy.schwartz1\}@mail.huji.ac.il}\\
  {\tt haop@allenai.org} \quad   {\tt \{jkasai,nasmith\}@cs.washington.edu}
  \quad {\tt ivamon@apple.com}
}

\begin{document}
\maketitle
\begin{abstract}
\input{0_abstract.tex}
\end{abstract}

\section{Introduction}
\label{sec:intro}
\input{1_intro.tex}

\section{Background: Attention in Transformers}
\label{sec:background}
\input{2_background.tex}

\section{The \methodName Method}
\label{sec:method}
\input{3_method.tex}

\section{Experiments}
\label{sec:experiments}
\input{4_experiments.tex}

\subsection{Probing Results}
\label{ssec:results}
\input{5_results.tex}

\section{Further Analysis}
\label{sec:ablation}
\input{6_analysis.tex}

\section{Related Work}
\label{sec:related_work}
\input{7_related_work.tex}

\section{Conclusion}
\label{sec:conclusion}
\input{8_conclusion.tex}

\section{Limitations}
\label{sec:limitations}
\input{9_limitations.tex}

\section*{Acknowledgments}
We thank Miri Varshavsky for the great feedback and moral support.
This work was supported in part
by NSF-BSF grant 2020793,
NSF grant 2113530, an Ulman Fellowship, a Google Fellowship, a Leibnitz Fellowship, and a research gift from Intel.

\bibliography{custom}
\bibliographystyle{acl_natbib}
\clearpage
\appendix

\section{Pre-Processing}
\label{app:pre_process}

\input{app_pre_processing}

\section{Hyperparameters}
\label{app:hyperparamters}
\input{app_hyper_parameters}

\section{Further Analsys results for \( \text{RoBERTa}_{\text{LARGE}} \)}
\label{app:large_analsys}
\input{app_large_analsys.tex}

\end{document}

%% file: 0_abstract.tex
{

The attention mechanism is considered the backbone of the widely-used Transformer architecture. It contextualizes the input by computing input-specific attention matrices.
We find that this mechanism, while powerful and elegant, is not as important as typically thought for pretrained language models.
We introduce \methodName,\footnote{\methodName stands for \textit{\textbf{P}robing \textbf{A}nalysis for \textbf{P}LMs' \textbf{A}ttention}.} a new probing method that replaces the input-dependent attention matrices with constant ones---the average attention weights over multiple inputs.
We use \methodName to analyze several established pretrained Transformers on six downstream tasks. 
We find that \emph{without} any input-dependent attention, all models achieve competitive performance---an average relative drop of only 8\% from the probing baseline.
Further, little or no performance drop is observed when replacing half of the input-dependent attention matrices with constant (input-\textit{independent}) ones.
Interestingly, we show that better-performing models lose more from applying our method than weaker models, suggesting that the utilization of the input-dependent attention mechanism might be a factor in their success.
Our results motivate research on simpler alternatives to input-dependent attention, as well as on methods for better utilization of this mechanism in the Transformer architecture.
}

%% file: 1_intro.tex
{

Pretrained Transformer \cite{Vaswani2017AttentionIA} models have enabled great progress in NLP in recent years
\cite{devlins2019bert,Liu2019RoBERTaAR,2020t5,gpt3, palm}.
A common belief is that the backbone of the Transformer model---and pretrained language models (PLMs) in particular---is the \textit{attention mechanism}, which applies multiple attention heads in 
parallel, each generating an \textit{input-dependent} attention weight matrix.

Interestingly, recent work found that attention patterns tend to focus on constant (input-\textit{independent}) positions \cite{clark-etal-2019-bert, voita-etal-2019}, while other works showed that it is possible to pretrain language models where the attention matrices are replaced with constant matrices without major loss in performance \cite{gmlp, fnet, flash}.
A natural question that follows is how much standard PLMs, pretrained \textit{with} the attention mechanism, actually rely on this input-dependent property.
This paper shows that they are less dependent on it than previously thought.

\begin{figure}[t]
\centering
\includegraphics[trim={0 1.5cm 0 0}, clip, width=\columnwidth]{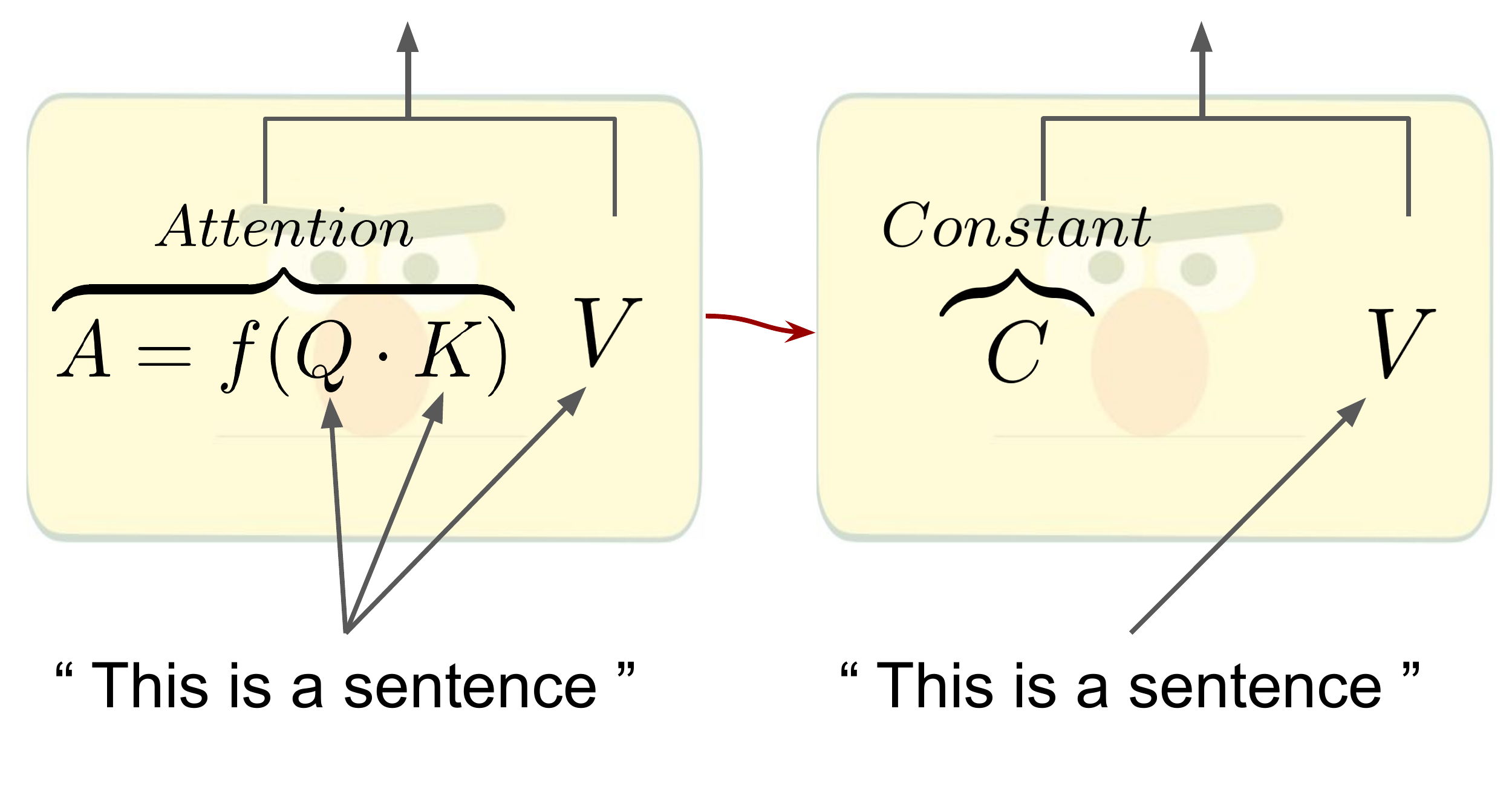}

\caption{\label{fig:teaser_1}Illustration of the \methodName method, which measures how much PLMs use the attention mechanism. \methodName replaces the input-dependent attention matrices (left) with constant ones (right). We then measure the performance gap between the two.  Moderate drop indicates minor reliance on the attention mechanism.
}
\end{figure}

We present a new analysis method for PLMs: Probing Analysis for PLMs’
Attention (\methodName).
For each attention head $h$, \methodName replaces the attention matrix with a constant one:
a simple average of the attention matrices for $h$ computed on some unlabeled corpus.
Replacing all attention matrices with such constant matrices results in an attention-free variant of the original PLM (See \figref{teaser_1}).
We then compute, for some downstream tasks, the probing performance gap between an \base model and its attention-free variant. 

This provides a tool to quantify the models' reliance on attention. 
Intuitively, a larger performance drop indicates that the model relies more on the input-dependent attention mechanism.

We use \methodName to study three established pretrained Transformers: BERT~\cite{devlins2019bert}, RoBERTa~\cite{Liu2019RoBERTaAR}, and DeBERTa~\cite{deberta}, each with BASE- and LARGE-sized versions. We evaluate these models on six diverse benchmarks, spanning text classification and structured prediction tasks.

Our results suggest that attention is not as important to pretrained Transformers as previously thought.
First, the performance of the attention-free variants is comparable to \base models: an average \emph{relative} drop of only 8\%.
Second, replacing half of the attention matrices with constant ones has little effect on performance, and in some cases may even lead to performance improvements.
Interestingly, our results hint that better models use their attention capability more than weaker ones; when comparing the effect of \methodName on different models, we find that the better the model's original performance is, the more it suffers from replacing the attention matrices with constant ones. This suggests a potential explanation for the source of the empirical superiority of some models over others---they make better use of the attention mechanism.

This work grants a better understanding of the attention mechanism in pretrained Transformers.
It also motivates further research on simpler or more efficient Transformer models, either for pretraining \cite{fnet, gmlp, flash} or potentially as an adaptation of existing pretrained models \cite{RFA,peng-etal-2022-abc,kasai2021t2r}. It also provides a potential path to improve the Transformer architecture---by designing inductive bias mechanisms for better utilization of attention  \cite{peng-etal-2020-mixture, Wang2022paying}.

Finally, our work may contribute to the ``attention as explanation'' debate \cite{jain-wallace-2019-attention,serrano-smith-2019-attention,wiegreffe-pinter-2019-attention,bibal-etal-2022-attention}.
By showing that some PLMs can perform reasonably well with constant matrices, we suggest that explanations arising from the attention matrices might not be crucial for models' success.

We summarize our main contributions. (1) We present a novel probing method---\methodName---which quantifies the reliance of a given PLM on its attention mechanism by ``disabling'' that mechanism for this PLM.
  (2) We apply  \methodName to six leading PLMs, and find that our manipulation leads to modest performance drops on average, which hints that attention might not be as important  as thought. (3) We show that better-performing PLMs tend to suffer more from our manipulation, which suggests that the input-dependent attention is a factor in their success. (4) Finally, we release our code and experimental results.\footnote{\url{https://github.com/schwartz-lab-NLP/papa}}

}

%% file: 2_background.tex
{

Transformers consist of interleaving attention and feed-forward layers.
In this work, we focus on Transformer encoder models, such as BERT, which are commonly used in many NLP applications. 

The (multi headed) self-attention module takes as input a matrix \(X \in \R^{n \times d}\) and produces a matrix \(X^{out} \in \R^{n \times d}\), where \(n\) denotes the number of input tokens, each represented as a \(d\)-dimensional vector. Each attention layer consists of \(H\) heads, and each head \(h \in \{1,\dots,H\}\) has three learnable matrices: \(W_Q^h, W_K^h, W_V^h \in \R^{d \times d'}\).\footnote{\(d'\) is the head-dimension, and usually defined as \(d'=\frac{d}{H}\).} Multiplying them with the input \(X\) results in: \(Q^h, K^h, V^h \in \R^{n \times d'}\) (the queries, keys and values, respectively).

The queries and the keys compute a $n\times n$ attention weight matrix \(A^h\) between all pairs of tokens as softmax-normalized dot products:\footnote{Some attention variants (e.g., \citealp{deberta}) incorporate positional information as part of the calculation of \(A^h\).}
\[
A^h=\text{softmax}\Bigl(\underset{Q^h}{\underbrace{\bigl(X \cdot W_Q^h\bigr)}} \cdot \bigl(\underset{K^h}{\underbrace{ X \cdot W_K^h}}\bigr)^\top\Bigr) \in \R^{n \times n}  \addtag
\label{A_matrix}
\]
where the softmax operation is taken row-wise.
The value matrix \(V^h\) is then left-multiplied by the attention matrix \(A^h\) to generate the attention head output.

Importantly, the attention matrix \(A^h\) is \textit{input-dependent}, i.e., defined by  the input \(X\). This property is considered to be the backbone of the attention mechanism \cite{Vaswani2017AttentionIA}.

An intriguing question is the extent to which PLMs actually rely on the attention mechanism. In the following, we study this question by replacing the attention matrices of PLMs with constant matrices. We hypothesize that if models make heavy use of attention, we will see a large drop in performance when preventing the model from using it. As shown below, such performance drop is often \emph{not} observed.

}

%% file: 3_method.tex
{
We present \methodName, a probing method for quantifying the extent to which pretrained Transformer models use the attention mechanism.  
\methodName works by replacing the Transformer attention weights with constant matrices, computed by averaging the values of the attention matrices over unlabeled inputs (\subsecref{replace_att}).
\methodName also allows for replacing any subset (not just all) of the attention matrices.
We propose a method for selecting which heads to replace (\subsecref{sorting_method}).
The resulting model is then probed against different downstream tasks (\subsecref{classifier}). The performance difference between the \base and the new models can be seen as an indication of how much the model uses its attention mechanism.

\subsection{Generating Constant Matrices}
\label{ssec:replace_att}
To estimate how much a pretrained Transformer \(m\) uses the attention mechanism, we replace its attention matrices with a set of constant ones, one for each head. To do so, \methodName constructs, for a given head $h$,\footnote{We do so for all layers in parallel.
Layer indices omitted for simplicity.} a constant matrix \(C^h\) by averaging the attention matrix  \(A^h\) over a corpus of raw text.
More specifically, given a corpus \(D = \{e_1, \dots, e_{|D|}\}\), \(C^h\) is defined as:  
\[C^h = \frac{1}{|D|} \sum_{i=1}^{|D|} A^h_i, \addtag  \label{B_matrix}\]
where \(A^h_i\) is the input-dependent attention matrix that \(h\) constructs while processing \(e_i\). We note that the average is taken entry-wise, and only over non-padded entries (padding tokens are ingored).

We emphasize that the construction process of \(C^h\) matrices requires no labels.
In \subsecref{ablAttMat} we compare our method of constructing constant matrices from unlabeled data to other alternatives that either use no data at all, or use labeled data.

\subsection{Replacing a Subset of the Heads}
\label{ssec:sorting_method}

Different attention heads may have different levels of dependence on attention. 
We therefore study the effect of replacing a subset of the heads, and keeping the rest intact. To do so, we would like to estimate the reliance of each head on the input-dependent attention, which would allow replacing only the heads that are least input-dependent for the model.

To estimate this dependence, we introduce a new weighting parameter \( \lambda^h \in (0, 1)\), initialized as \(\lambda^h = 0.5 \), for each attention head \(h\).\footnote{\(\lambda^h\) is the output of a sigmoid over a learned parameter.
}
\(\lambda^h\) is a learned weighting of the two matrices: the attention matrix \(A^h\) and the constant matrix \(C^h\) from \eqref{A_matrix} and \eqref{B_matrix} respectively. For each input \(e_i\), a new matrix \(B^h\) is constructed as:
\[ B^h_i = \lambda^h \cdot A^h_i + (1 - \lambda^h) \cdot C^h \addtag \]
We interpret a smaller \(\lambda^h\) as an indication of $h$ less depending on the attention mechanism.

We then train the probing classifier (\subsecref{classifier}) along with the additional \( \lambda^h \) parameters.
We use the learned \(\lambda^h\)s to decide which heads should be replaced with constant matrices, by only replacing the \(k\%\) attention heads with the  smallest \( \lambda^h \) values for some hyperparameter \(k\).\footnote{In \subsecref{ablSort} we show that this head selection method outperforms other alternatives.} Importantly, this procedure is only used as a pre-processing step; our experiments are trained and evaluated without it, where \(k\%\) of each model's heads are replaced, and \((1-k\%)\) remain unchanged.

\subsection{Probing}
\label{ssec:classifier}
Our goal is to evaluate how much attention a given PLM uses. Therefore, we want to avoid finetuning it for a specific downstream task, as this would lead to changing all of its weights, and arguably answer a different question (e.g., how much attention does a \textit{task-finetuned} PLM use). Instead, we use a probing approach~\cite{probing_nelson, probing_belinkov} by freezing the model and adding a classifier on top.

Our classifier calculates for each layer a weighted (learned, non-attentive) representation of the different token representations.
It then concatenates the different layer weighted representations, and applies a 2-layer MLP. For structured prediction tasks (e.g., NER and POS), {where a representation for each token is needed}, 
we concatenate for each token the representations across layers, and apply a 2-layer MLP.

When \methodName is applied to some input, we replace the attention matrices \(A^h\) with the corresponding constant matrices \(C^h\).\footnote{To minimize model changes, we also mask the \(C^h\) entries corresponding to padded tokens, and normalize the matrix (row-wise), as in a regular Transformer.}
We then compare the downstream performance of the \base model \(m\) with the new model \(m'\). 
The larger the performance gap between \(m\) and \(m'\), the higher \(m\)'s dependence on the attention mechanism.

\subsection{Method Discussion}
\label{ssec:method_discussion}
\paragraph{Contextualization with \methodName}
\methodName replaces the attention matrices with constant ones, which results in an attention-free model. Importantly, unlike a feed-forward network, the representations computed via the resulting model are still contextualized, i.e., the representation of each word depends on the representations of all other words. The key difference between the standard Transformer model and our attention-free model is that in the former the contextualization varies by the input, and for the latter it remains fixed for all inputs.

\paragraph{Potential Computational Gains}The replacement of the attention matrix with a constant one motivates the search for 
efficient attention alternatives.
Using constant matrices is indeed more efficient, reducing the attention head time complexity from \(2n^2d' + 3nd'^2\) to \(n^2d' + nd'^2\),\footnote{\(n\) is the sequence length and \(d'\) is head-dimension.} which shows potential for efficiency improvement.

Several works used various approaches for replacing the attention mechanism with constant ones during the pretraining phase \cite{fnet,gmlp,flash}, and indeed some of them showed high computational gains.
Our work tackles a different question---how much do PLMs, which trained with the attention mechanism, actually use it.
Thus, unlike the approaches above,
we choose to make minimal changes to the \base models. Nonetheless, our results further motivate the search for efficient attention variants.

}

%% file: 4_experiments.tex
{
We now turn to use \methodName to study the attention usage of various PLMs.

\subsection{Experimental Setup}
\label{ssec:experimental_setup}

Our experiments are conducted over both text classification and structured prediction tasks, all in English. For the former we use four diverse benchmarks from the GLUE benchmark \cite{GLUE}: MNLI \cite{williams-etal-2018-broad}, SST-2 \cite{socher-etal-2013-recursive}, MRPC \cite{dolan-brockett-2005-automatically}, and CoLA \cite{warstadt-etal-2019-neural}. For the latter we use named entity recognition (NER) and part of speech tagging (POS) from the CoNLL-2003 shared task \cite{tjong-kim-sang-de-meulder-2003-introduction}.\footnote{We report accuracy for SST2 and MNLI, F1 score for MRPC, NER and POS, and MCC for CoLA. 
 }
We use the standard train/validation splits, and report validation results in all cases.\footnote{For MNLI, we report the mismatched validation split.}

We use three widely-used pretrained Transformer encoder models: BERT \cite{devlins2019bert}, RoBERTa \cite{Liu2019RoBERTaAR}, and DeBERTa \cite{deberta}.
We use both BASE (12 layers, 12 heads in each layer) and LARGE (24 layers, 16 heads per layer) versions.
For each model and each task, we generate the constant matrices with the given (unlabeled) training set of that task. In \subsecref{staticIndependence} we show that \methodName is not very sensitive to the specific training set being used.

All experiments are done with three different random seeds, average result is reported (95\% confidence intervals are shown).
Pre-processing and additional experimental details are described in \appref{pre_process} and \ref{app:hyperparamters}, respectively.

}

%% file: 5_results.tex
{

\begin{figure*}[htp]
\centering
\begin{subfigure}[htp]{\textwidth}
\includegraphics[width=\textwidth, trim={0 1.25cm 0 0},clip]{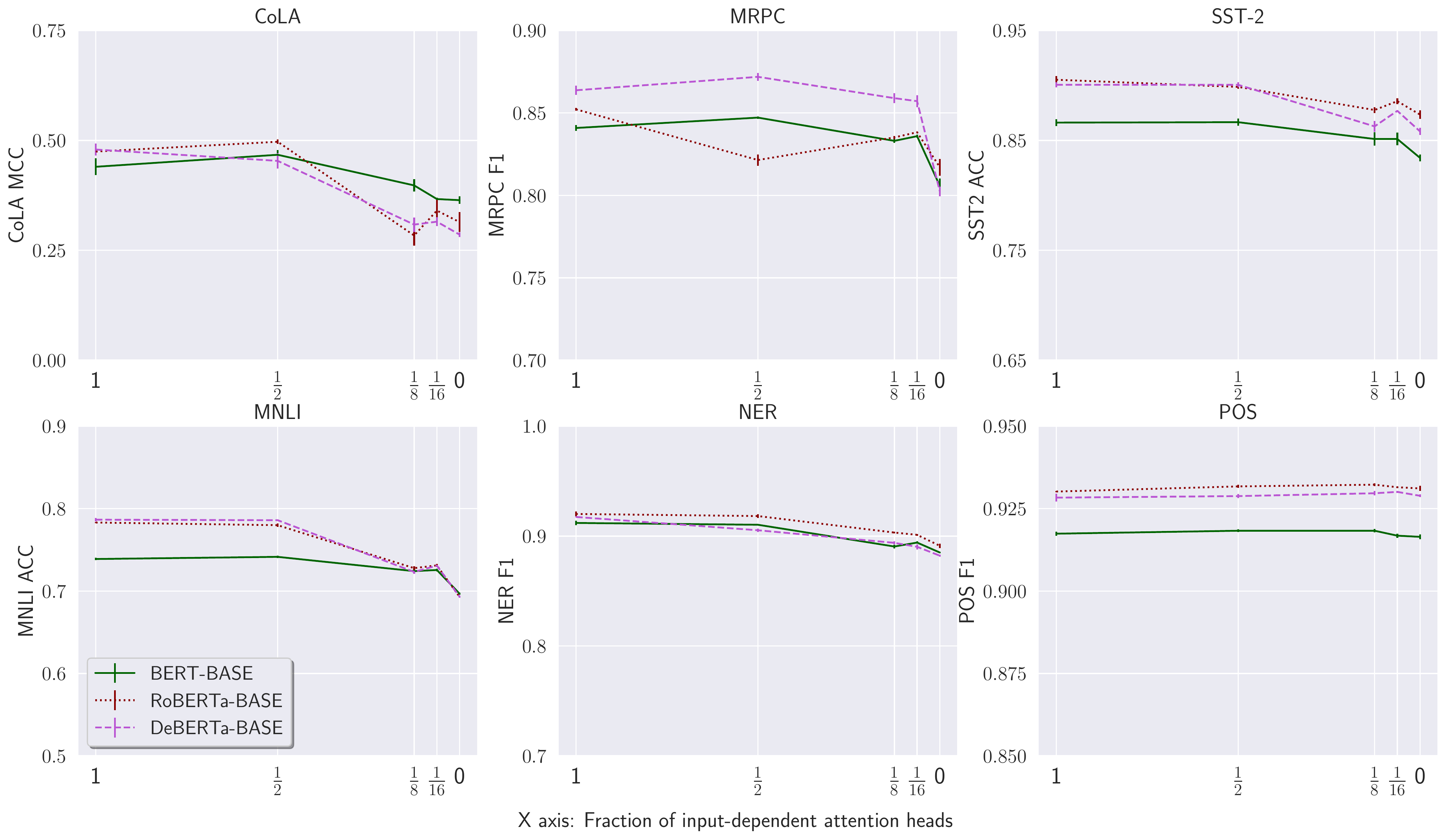}
\caption{BASE models \label{fig:base_models}}
    \end{subfigure}
\begin{subfigure}[htp]{\textwidth}
\includegraphics[width=\textwidth]{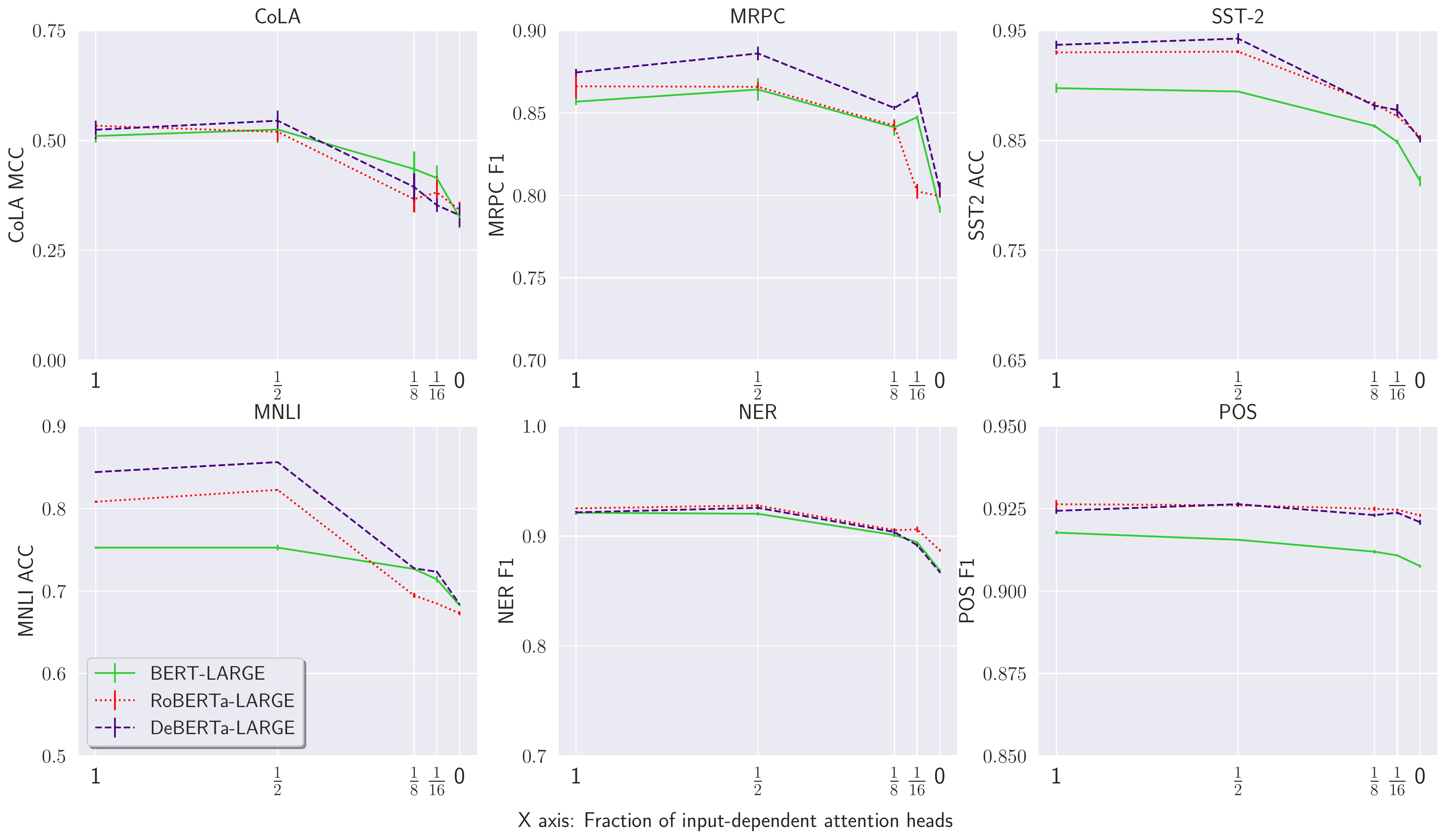}
\caption{LARGE models\label{fig:large_models}}
\end{subfigure}

\caption{\label{fig:results}Probing results ($y$-axis) with decreasing number of attention heads ($x$-axis). BASE models are shown in \figref{base_models}, and LARGE models are shown in \figref{large_models}. Higher is better in all cases.%
}
\end{figure*}

The results of the BASE and LARGE models are presented in \figref{base_models} and \ref{fig:large_models}, respectively. We measure the performance of each model on each task  using  \( \{ 1, \frac{1}{2}, \frac{1}{8}, \frac{1}{16} ,0 \} \) of the model's input-dependent attention matrices and replacing the rest with constant ones.

We first consider the original, fully-attentive, models, and find that performance decreases in the order of DeBERTa, RoBERTa, and BERT.

This order is roughly maintained across tasks and model sizes, which conforms with previous results of fine-tuning these PLMs \cite{deberta}.
This suggests that the model ranking of our probing method is consistent with the standard fine-tuning setup.

We note that the trends across tasks and models are similar; hence we discuss them all together in the following (up to specific exceptions).

\paragraph{Replacing \emph{all} attention matrices with constant ones incurs a moderate performance drop} 
As shown in \figref{results}, applying \methodName on all attention heads leads to an 8\% relative performance drop on average and not greater than 20\% from the \base model.\footnote{For the MRPC task, some of the attention-free models do get close to the majority baseline, though still above it.} This result suggests that pretrained models only moderately rely on the attention mechanism.

\paragraph{Half of the attention matrices can be replaced without loss in performance} We note that in almost all cases replacing half of the models' attention matrices leads to no major drop in performance. In fact, in some cases, performance even \emph{improves} compared to the \base model (e.g., \(\text{BERT}_{\text{BASE}}\) and \(\text{DeBERTa}_{\text{LARGE}}\)), suggesting that some of the models' heads have a slight preference towards constant matrices.
This result is consistent with some of the findings of recent hybrid models that use both constant and regular attention \cite{gmlp,fnet} to build efficient models.

\paragraph{Performant models rely more on attention}
\begin{figure}[t]
\includegraphics[width=\columnwidth]{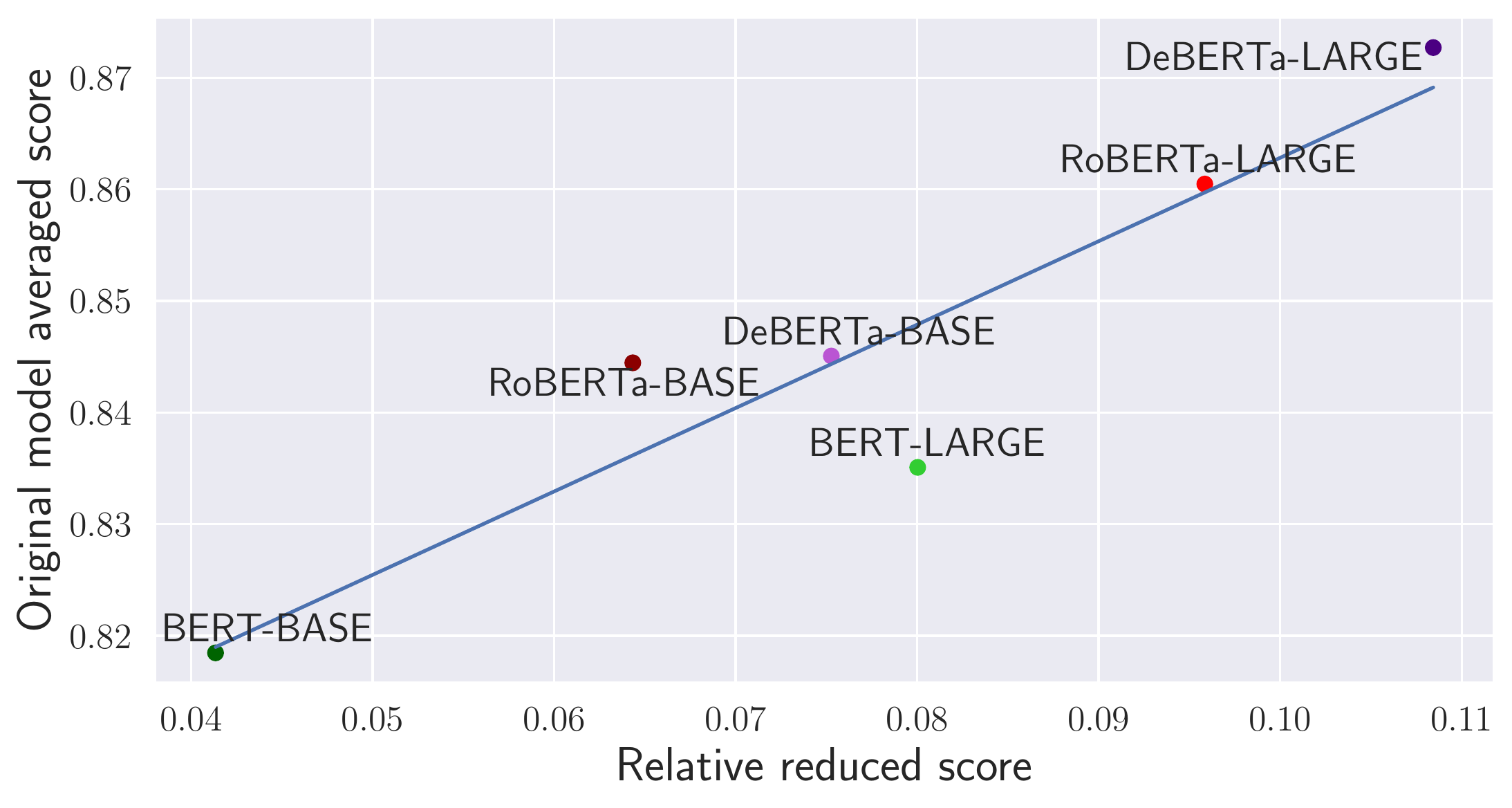}
\centering
\caption{\label{fig:perfomance_vs_reduction}
Stronger-performing PLMs use their attention capability more. \(y\)-axis: \base model average performance; \(x\)-axis: relative reduced score  when all attention matrices are replaced with constant ones.}
\end{figure}

\figref{perfomance_vs_reduction} shows for each model the relation between the \base performance (averaged across tasks) and the averaged (relative) reduced score when replacing all attention heads. We observe a clear trend between the models' performance and their relative reduced score, which suggests that better performing models use their attention mechanism more.

}

%% file: 6_analysis.tex
{

We present an analysis of \methodName, to better understand its properties. We first discuss the patterns of the constant matrices produced by \methodName (\subsecref{patterns}). Next, we consider other alternatives to generating constant matrices (\subsecref{ablAttMat}); we then examine whether the constant matrices are data-dependent (\subsecref{staticIndependence}); we continue by exploring alternative methods for selecting which attention heads to replace (\subsecref{ablSort}). Finally, we  present MLM results, and discuss the challenges in interpreting them (\subsecref{MLMexp}).
In all experiments below, we use \( \text{RoBERTa}_{\text{BASE}}\). \( \text{RoBERTa}_{\text{LARGE}}\) experiments  show very similar trends, see~\appref{large_analsys}.

\subsection{Patterns of the Constant Matrices}
\label{ssec:patterns}
\begin{figure*}[t]
\includegraphics[trim={0 6.6cm 0 0},clip,width=\textwidth]{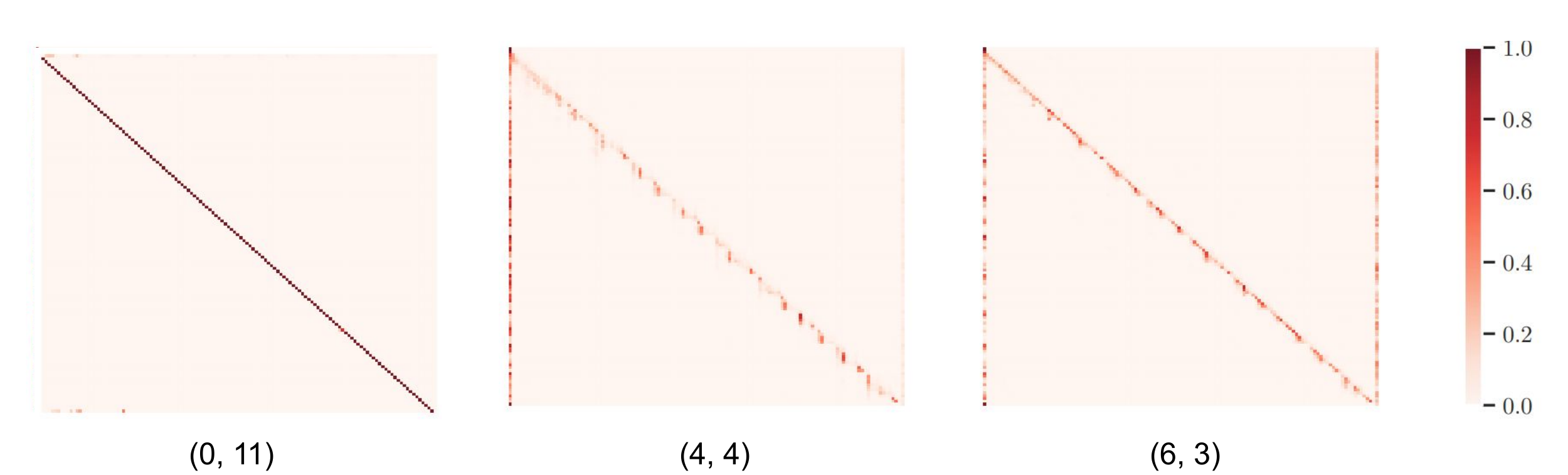}
\centering
\caption{\label{fig:heatmaps}Generated constant matrices  \(C^h\) by the \methodName method for representative heads (layer, head). These matrices used for the attention-free variant of \( \text{RoBERTa}_{\text{BASE}} \) for the SST-2 task.}
\end{figure*}

We first explore the attention patterns captured by different heads by observing the constant matrices (\(C^h\)). We first notice a  diagonal pattern, in which each token mostly attends to itself or to its neighboring words. This pattern is observed in about 90\% of the constant matrices produced by \methodName. Second, about 40\% of the heads put most of their weight mass on the [CLS] and/or [SEP] tokens (perhaps in combination with the diagonal pattern described above). Lastly, while for some of the heads the weight mass is concentrated only in specific entry per row (which corresponding only to a specific token), in most of cases the weight mass is distributed over several entries (corresponding to several different tokens). These patterns are similar to those identified by \citet{clark-etal-2019-bert}, and explain in part our findings---many of the attention heads mostly focus on fixed patterns that can also be captured by a constant matrix. \figref{heatmaps} shows three representative attention heads that illustrate the patterns above.

\subsection{Alternative Constant Matrices}
\label{ssec:ablAttMat}
\input{tables/matrix_ablation}
\methodName replaces the attention matrices with constant ones. As described in \subsecref{replace_att}, this procedure requires only an unlabeled corpus. In this section, we compare this choice with constant matrices that are constructed without any data (data-free matrices), and those that require labeled data for construction (labeled matrices).

For the former we consider three types of matrices: (1) Identity matrix---in which each token `attends' only to itself, and essentially makes self-attention a regular feed-forward (each token is processed separately); (2) Toeplitz matrix---we use a simple Toeplitz matrix (as suggested in \citealp{gmlp}), where the weight mass is on the current token, and it decreases as the attended token is further from the current one (the entries of the matrix are based on the harmonic series);\footnote{Similar to the Gaussian matrices suggested by \citet{gauss_mt}.} (3) Zeros matrix---essentially pruning the heads.

We also consider two types of labeled-matrices: (4) initialized as the Toeplitz matrices from (2); and (5) initialized as our average matrices. These matrices are updated during the training procedure of the probing classifier.\footnote{To make minimal changes to the frozen model, all constant matrices are masked and normalized (row-wise), the same as the output of the original softmax operation.}

\tabref{matrix_abl} shows the performance of each attention-free resulting model for all downstream tasks. We observe that for all tasks, our average-based model outperforms all other data-free models by a notable margin. 
As for the labeled-matrices models, our model also outperforms the one initialized with Toeplitz matrices (4), and in most cases gets similar results to the model initialized with average matrices (5).
 It should be noted that the \base models (with regular attention) do not update their inner parameters in the probing training phase, which makes the comparison to the labeled-matrices models somewhat unfair. The above suggests that our choice of constant matrix replacement better estimates the performance of the attention-free PLMs.

\subsection{Are the Constant Matrices Data-Dependent?}
\label{ssec:staticIndependence}

\input{tables/mnli_avgs}

\methodName constructs the constant matrix for a given head \(C^h\) as the average of the model's attention matrices over a given corpus \(D\), which in our experiments is set to be the training set of the task at hand (labels are not used). 
Here we examine the importance of this experimental choice by generating \(C^h\) using a different dataset---the MNLI training set, which is out of distribution for the other tasks.

Results are presented in \tabref{mnli_avgs}. The performance across all tasks is remarkably similar between generating the matrices using the specific task training set and MNLI, which suggests that the constant matrices might be somewhat data-independent.

\subsection{Alternative Head Selection Methods}
\label{ssec:ablSort}
We compare our method for selecting which heads to replace (\subsecref{sorting_method}) with a few alternatives. The first two replace the heads by layer order: (1) we sort the heads from the model's first layer to the last and (2) from the model's last layer to the first. In both cases we use the internal head ordering per layer for ordering within the layer. We then replace the first \(k\%\) of the heads. We also add (3) a random baseline that randomly replaces \(k\%\) of the heads, and a (4) `Reversed' one which replaces the heads with the \textit{highest} (rather than lowest) \(\lambda^h\) values (\subsecref{sorting_method}). 

\begin{figure}[t]
\includegraphics[width=\columnwidth]{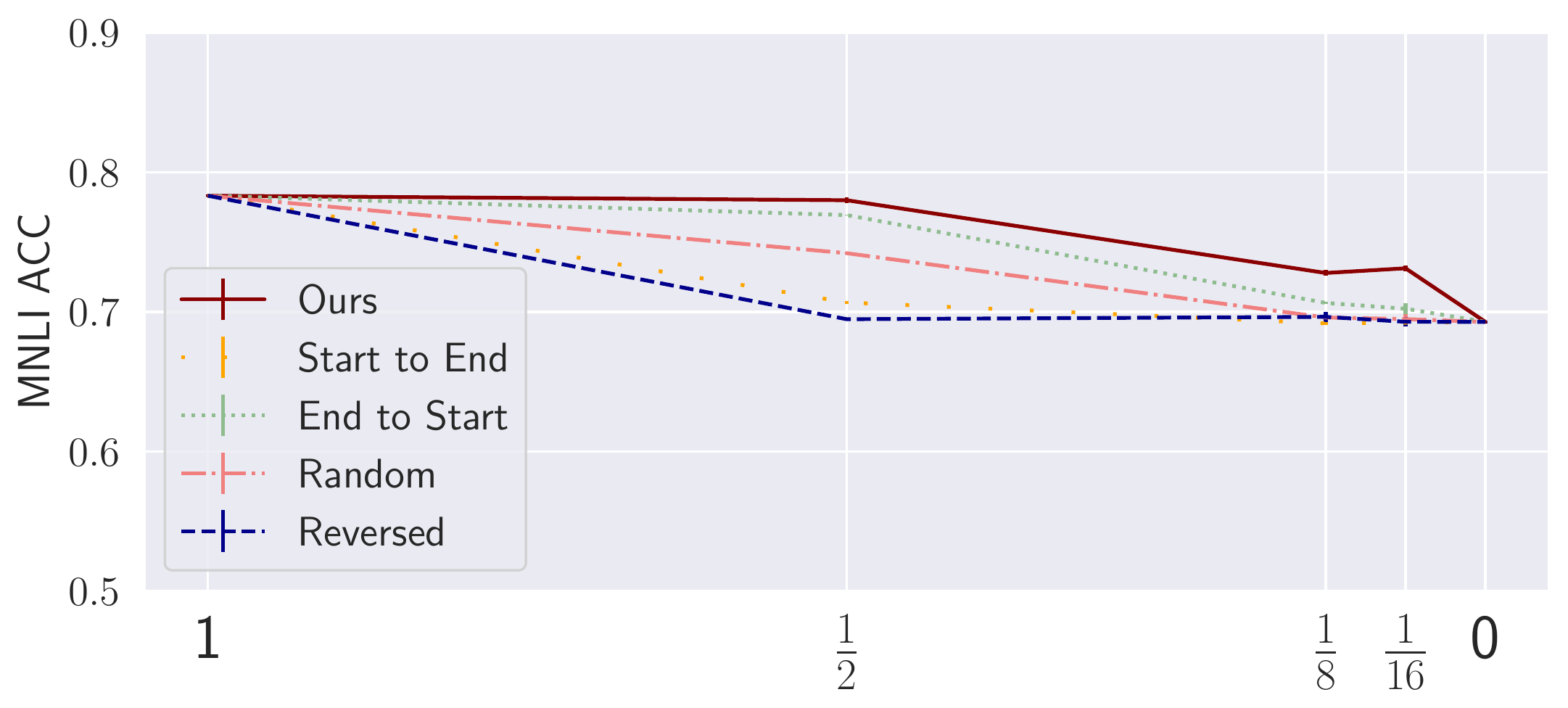}
\centering
\caption{\label{fig:sort_ablation}Comparison between different heads selection methods over MNLI. Our method outperforms all other alternatives. The x-axis represents the fraction of input-dependent attention heads.}
\end{figure}

\figref{sort_ablation} shows the MNLI performance of each method as a function of the fraction of heads replaced. We observe that our method, which is based on learned estimation of attention importance, outperforms all other methods for every fraction of heads replaced. Moreover, the `Reversed' method is the worst among the examined methods, which suggests that our method not only replaces the least attention dependent heads first, but also replaces the most dependent ones last.
 Although our head replacement order outperforms the above methods, we note that our order is an overestimation of the model attention dependency, and better methods might show that even less attention is needed.

\subsection{Effects on MLM Perplexity}
\label{ssec:MLMexp}
\begin{figure*}[th]
    \centering
    \centering
    \begin{subfigure}[tb]{0.495\textwidth}
    \includegraphics[trim={0 -0.7cm 0 0},clip,width=\textwidth]{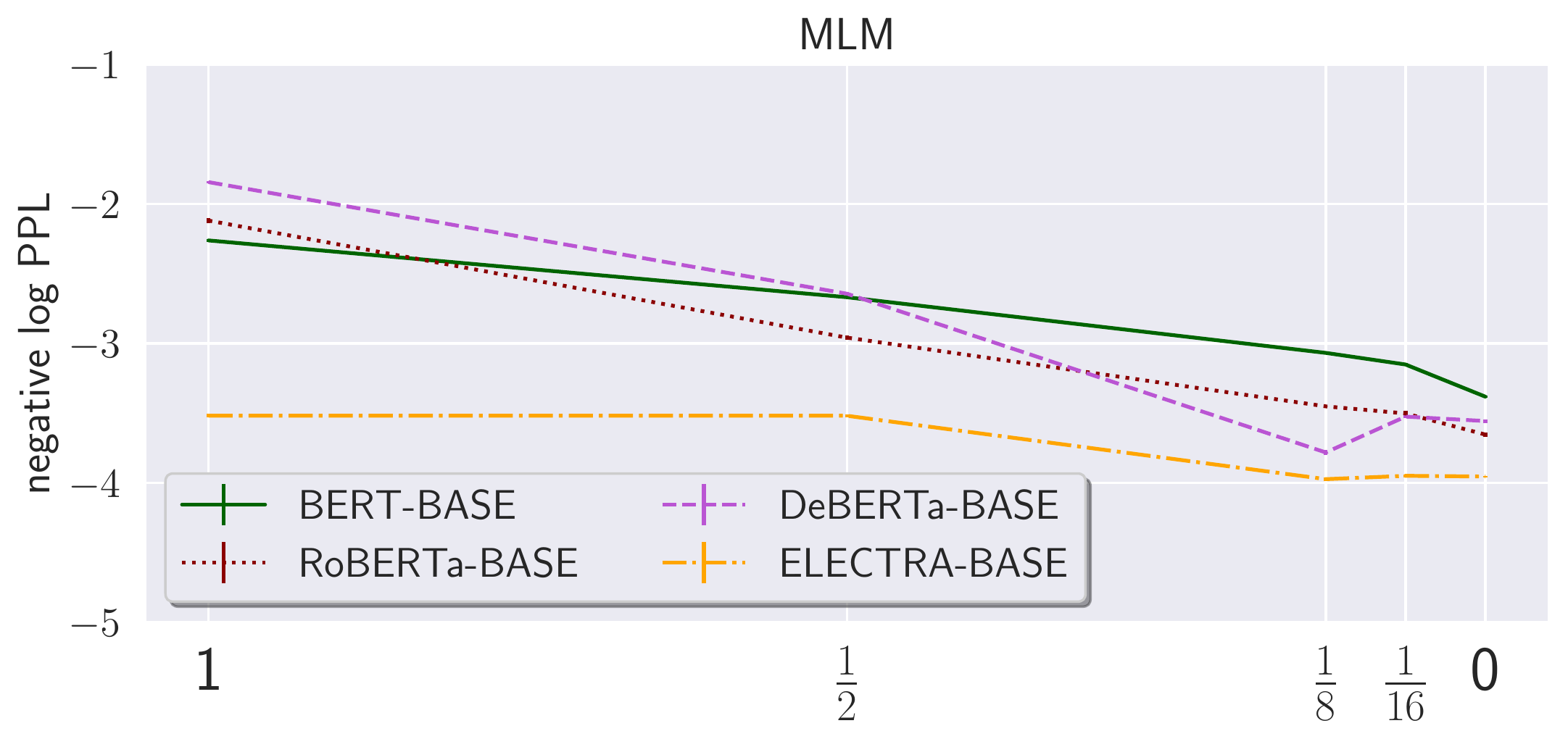}
    \caption{\label{fig:mlm_for_mlm}MLM results}
    \end{subfigure}
    \begin{subfigure}[tb]{0.495\textwidth}
    \includegraphics[trim={0 -0.7cm 0 0},clip,width=\textwidth]{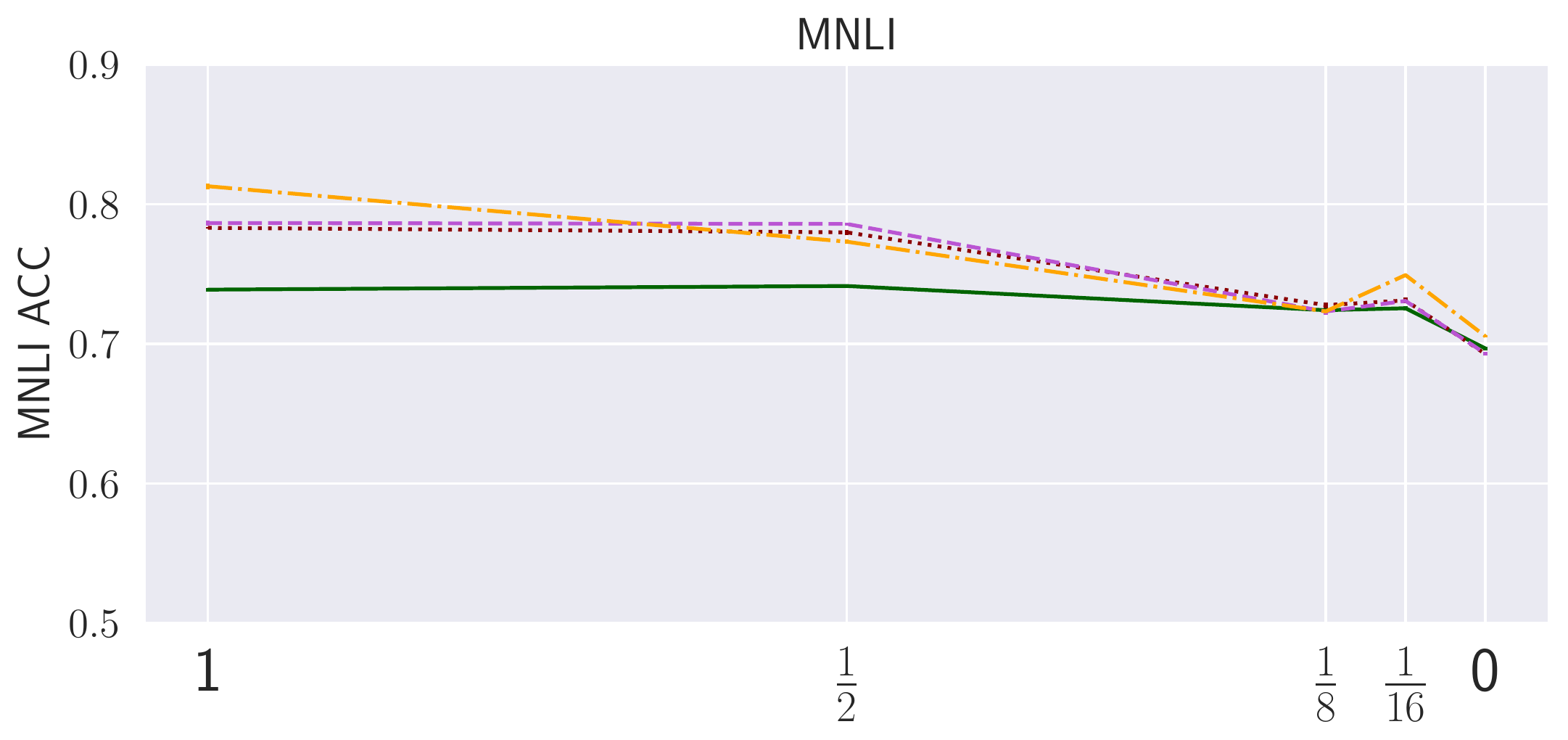}
    \caption{\label{fig:mnli_for_mlm}MNLI mismatched results}
    \end{subfigure}
    \caption{\label{fig:mlm} ELECTRA\textsubscript{BASE} model compared with other BASE models on MLM and MNLI. In \figref{mlm_for_mlm} ELECTRA\textsubscript{BASE} behaves similarly to its behavior on MNLI, but not to the other models, which are MLM-based. In \figref{mnli_for_mlm} ELECTRA\textsubscript{BASE} behaves similar to other models. In both graphs the \(x\)-axis represents the fraction of input-dependent attention heads, and the \(y\)-axis is the score of the specific task (higher is better).}
\end{figure*}
So far we have shown that applying \methodName on downstream tasks only incurs a moderate accuracy drop.
This section aims to explore its impact on masked language modeling (MLM).
We find that while our models suffer a larger performance drop on this task compared to the other tasks, this can be explained by their pretraining procedure.

\figref{mlm_for_mlm} plots the negative log perplexity (higher is better) of all BASE models on the WikiText-103  \cite{wiki103} validation set.
When replacing attention matrices using \methodName,
MLM suffers a larger performance drop compared to the downstream tasks (\subsecref{results}).
We hypothesize that this is because these pretrained Transformers are more specialized in MLM, the task they are pretrained on.
As a result, they are less able to adapt to architectural changes in MLM than in downstream tasks.
To test our hypothesis, we probe ELECTRA\textsubscript{BASE} \cite{clark2020electra} using \methodName.
ELECTRA is an established pretrained Transformer trained with the \emph{replaced token detection} objective, instead of MLM.
It has proven successful on a variety of downstream tasks.

ELECTRA\textsubscript{BASE}'s probing performance on MLM supports our hypothesis: We first note that its \base performance is much worse compared to the other models (--3.51 compared to around --2 for the MLM-based models), despite showing similar performance on downstream tasks (\figref{mnli_for_mlm}), which hints that this model is much less adapted to MLM. Moreover, the drop when gradually removing heads is more modest (a 0.44 drop compared to 1--1.5 for the other models), and looks more similar to ELECTRA\textsubscript{BASE}'s probing performance on MNLI (\figref{mnli_for_mlm}).
Our results suggest a potential explanation for the fact that some pretrained Transformers suffer a larger performance drop on MLM than on downstream tasks;
rather than MLM demanding higher attention use, this is likely because these models are pretrained with the MLM objective.

}

%% file: tables/matrix_ablation.tex
\begin{table*}[th]
\centering
\begin{tabular}{@{} cc|cccccc @{}}
\hline
\textbf{Matrix Construction} & \textbf{Matrix Type} & \textbf{CoLA} & \textbf{MRPC} & \textbf{SST2} & \textbf{MNLI-mm} & \textbf{NER} & \textbf{POS}\\
\hline
\hline
Attention based & Original & 0.47 & 0.85 & 0.91 & 0.78 & 0.92 & 0.93\\
\hline

\multirow{3}{6em}{Data-Free} 
& Identity  & 0.04 & 0.80 & 0.80 & 0.63 & 0.55 & 0.87 \\
& Toeplitz  & 0.08 & 0.81 & 0.79 & 0.65 & 0.77 & 0.90 \\
& Zeros   & 0.09 & 0.80 & 0.80 & 0.66 & 0.57 & 0.87 \\
\multirow{2}{6em}{Labeled Data} 
& Toeplitz init.   & 0.08 & 0.81 & 0.79 & 0.68 & 0.78 & 0.91 \\
& Average init.    & \textbf{0.34} & 0.81 & \textbf{0.87} & \textbf{0.72} & \textbf{0.89} & \textbf{0.93} \\
Unlabeled Data &Average \textbf{(Ours)} & 0.31 & \textbf{0.82} & \textbf{0.87} & 0.69 & \textbf{0.89} & \textbf{0.93} \\
\hline
\end{tabular}
\caption{\label{tab:matrix_abl}
Probe task of performance of \( \text{RoBERTa}_{\text{BASE}} \) with different constant matrix types as a replacement to the input-dependent attention matrix. Bold numbers indicate the best constant model for the task. Our approach based on an average of multiple attention matrices outperforms all other data-free matrix types across all tasks, and gets similar results to the best labeled-data based model. In all tasks higher is better. 
}
\end{table*}

%% file: tables/mnli_avgs.tex
\begin{table}[t]
\setlength{\tabcolsep}{5pt}
\centering
\begin{tabularx}{\columnwidth}{@{} c|ccccc@{}}
\hline
\textbf{Task} & \textbf{CoLA} & \textbf{MRPC} & \textbf{SST2} &  \textbf{NER} &  \textbf{POS}\\
\hline
\hline
Per-Task & 0.31 & 0.82 & 0.87 & 0.89 & 0.93 \\
MNLI & 0.32 & 0.81 & 0.87  & 0.89 & 0.93 \\
\hline
\end{tabularx}

\caption{\label{tab:mnli_avgs}
Comparison of probe task performance of \( \text{RoBERTa}_{\text{BASE}} \) between two setups of constructing the averaged constant matrices \(C^h\): Per-Task uses the task training set, while MNLI uses the constant matrices generated with the MNLI dataset.
The results are similar between the two setups, which indicates a low dependence of the constant matrices on the dataset used for constructing them.
}
\end{table}

%% file: 7_related_work.tex
{

\paragraph{Attention alternatives} Various efforts have been made in search of a simple or efficient alternative for the attention mechanism. Some works focused on building a Transformer variant based on an efficient approximation of the attention mechanism \cite{Kitaev2020ReformerTE,Wang2020LinformerSW,RFA,performer, schlag2021linear,cosformer}. Another line of research, which is more related to our work, replaced the attention mechanism in Transformers with a constant (and efficient) one. For instance, FNet \cite{fnet} replaced the attention matrix with the Vandermonde matrix, while gMLP \cite{gmlp} and FLASH \cite{flash} replaced it with a learned matrix.\footnote{These models also added a gating mechanism, which does not change the input-independent nature of their component.} These works showed that pretraining attention-free LMs can lead to competitive performance.
Our work shows that PLMs trained with attention can get competitive performance even if they are denied access to this mechanism during transfer learning.

\paragraph{Analysis of attention patterns}
Some investigations of how attention patterns in Transformers work use probing techniques. \citet{clark-etal-2019-bert}, \citet{Ravishankar2021AttentionCR} and \citet{Htut2019DoAH} studied the attention behavior in BERT. Unlike the above, which only focuses on the attention patterns of the PLM, our work sheds light on the dependence of PLMs on their attention mechanism.

\paragraph{Pruning methods} In this work we replaced the attention matrix with a constant one in order to measure the importance of the input-dependent ability. Works like \citet{Michel2019AreSH} and \citet{Li2021DifferentiableSP} pruned attention heads in order to measure their importance for the task examined. These works find that for some tasks, only a small number of unpruned attention heads is sufficient, and thus relate to the question of how much attention does a PLM use.
In this work we argue that replacing attention matrices with constant ones provides a more accurate answer for this question compared to pruning these matrices, and propose \methodName, a method for constructing such constant matrices.

}

%% file: 8_conclusion.tex
{

In this work, we found that PLMs are not as dependent on their attention mechanism as previously thought. To do so, we presented \methodName---a method for analyzing the attention usage in PLMs. We applied \methodName to several widely-used PLMs and six downstream tasks. Our results show that replacing all of the attention matrices with constant ones achieves competitive performance to the \base model, and that half of the attention matrices can be replaced without any loss in performance. We also show a clear relation between a PLM's aggregate performance across tasks and its degradation when replacing all attention matrices with constant ones, which hints that performant models make better use of their attention.

Our results motivate further work on novel Transformer architectures with more efficient attention mechanisms, both for pretraining and for knowledge distillation of existing PLMs.
They also motivate the development of Transformer variants that improve performance by making better use of the attention mechanism.
}

%% file: 9_limitations.tex
{
This work provides an analysis of the attention mechanism in PLMs. Our \methodName method is based on probing rather than finetuning, which is more common use to PLMs. We recognize that the attention mechanism in finetuned PLMs might act differently than the \base model, but our main focus is investigating the PLM itself, rather than its finetuned version.

Our analysis method is built on replacing the attention matrices with constant ones (\subsecref{replace_att}). We build these constant matrices by averaging the attention matrices over a given dataset. Because of this choice, our results reflect a lower bound on the results of the optimal attention-free model, and we acknowledge that there might be methods for constructing the constant matrices that would lead to even smaller gaps from the \base model. A similar argument can be applied for our heads selection method (\subsecref{sorting_method}). Importantly, better methods for these sub-tasks might further reduce the gap between the \base models and the attention-free ones, which will only strengthen our argument.

Finally, we note that we used the \methodName method with six English tasks, and recognize that results might be different for other tasks and other languages.

}

%% file: app_pre_processing.tex
To make the replacement of the attention matrix with a constant one reasonable, we fix the position of the [SEP] token to always be the last token of the model's input, rather than separating the last input token from the padding tokens (i.e., it comes after the padding tokens rather than before them). 
For tasks with two sequences per example (e.g., MNLI), which are typically separated by an additional [SEP] token, we fix this token to always be the middle token of the sequence, followed by the second sentence.
We recognize that this might lead to suboptimal usage of the input's sequence length, e.g., if one of the sentences is substantially longer than the other and particularly if it is longer than half of the sequence length, it would thus be trimmed. In our experiments this only happened in less than 0.2\% of input samples for a single task (MNLI), but we recognize that this might happen more frequently in other datasets.

%% file: app_hyper_parameters.tex
All of our code was implemented with the Transformers library \cite{wolf-etal-2020-transformers}. 
Hyperparameters for the probing classifier on downstream tasks are shown in \tabref{hyperparameters}.
\input{tables/hyperparamters}

%% file: tables/hyperparamters.tex
\begin{table}[!htbp] 
    \setlength{\tabcolsep}{4pt}
    \centering

\begin{tabular}{@{} c|cccc @{}}
& Learning Rate & Batch & Epochs & Seq.~Len.  \\\hline
CoLA & 2.00E-05 & 16 & 15 & 64 \\
SST-2 & 1.00E-04 & 32 & 4 & 64 \\
MNLI & 2.00E-04 & 8 & 4 & 256 \\
MRPC & 2.00E-05 & 16 & 15 & 128 \\
NER & 1.00E-04 & 8 & 4 & 128 \\
POS & 5.00E-04 & 8 & 4 & 128 \\
MLM & 5.00E-04 & 8 & 2 & 128 \\
\end{tabular}
    \caption{Probing classifier hyperparameters for downstream tasks.}
    \label{tab:hyperparameters}
\end{table}

%% file: app_large_analsys.tex
Tab.~\ref{tab:matrix_abl_large} and \ref{tab:mnli_avgs_large} show \( \text{RoBERTa}_{\text{LARGE}}\)'s analysis results for the experiments described in Sec.~\ref{ssec:ablAttMat} and \ref{ssec:staticIndependence}, respectively.

\input{tables/matrix_ablation_large}
\input{tables/mnli_avgs_large}

%% file: tables/matrix_ablation_large.tex
\begin{table*}[th]
\centering
\begin{tabular}{@{} cc|cccccc @{}}
\hline
\textbf{Matrix Construction} & \textbf{Matrix Type} & \textbf{CoLA} & \textbf{MRPC} & \textbf{SST2} & \textbf{MNLI-mm} & \textbf{NER} & \textbf{POS}\\
\hline
\hline
Attention based & Original & 0.53 & 0.87 & 0.93 & 0.81 & 0.93 & 0.93\\
\hline

\multirow{3}{6em}{Data-Free} 
& Identity  & 0.09 & 0.72 & 0.80 & 0.65 & 0.56 & 0.87 \\
& Toeplitz  & 0.11 & 0.79 & 0.80 & 0.65 & 0.74 & 0.89 \\
& Zeros   & 0.09 & 0.80 & 0.81 & 0.66 & 0.57 & 0.87 \\
\multirow{2}{6em}{Labeled Data} 
& Toeplitz init.   & 0.11 & 0.78 & 0.80 & 0.68 & 0.75 & 0.89 \\
& Average init.    & \textbf{0.35} & \textbf{0.81} & \textbf{0.88} & \textbf{0.73} & \textbf{0.91} & \textbf{0.93} \\
Unlabeled Data &Average \textbf{(Ours)} & 0.34 & \textbf{0.81} & 0.85 & 0.68 & 0.89 & 0.92 \\
\hline
\end{tabular}
\caption{\label{tab:matrix_abl_large}
Probe task of performance of \( \text{RoBERTa}_{\text{LARGE}} \) with different constant matrix types as a replacement to the input-dependent attention matrix. 
\tabref{matrix_abl} shows the results for \( \text{RoBERTa}_{\text{BASE}} \).
}
\end{table*}

%% file: tables/mnli_avgs_large.tex
\begin{table}[th]
\setlength{\tabcolsep}{5pt}
\centering
\begin{tabularx}{\columnwidth}{@{} c|ccccc@{}}
\hline
\textbf{Task} & \textbf{CoLA} & \textbf{MRPC} & \textbf{SST2} &  \textbf{NER} &  \textbf{POS}\\
\hline
\hline
Per-Task & 0.34 & 0.80 & 0.85 & 0.89 & 0.92 \\
MNLI & 0.35 & 0.81 & 0.85 & 0.88 & 0.92 \\
\hline
\end{tabularx}

\caption{\label{tab:mnli_avgs_large}
Comparison of probe task performance of \( \text{RoBERTa}_{\text{LARGE}} \) between two setups of constructing the averaged constant matrices \(C^h\): Per-Task uses the task training set, while MNLI uses the constant matrices generated with the MNLI dataset.
\tabref{mnli_avgs} shows the results for \( \text{RoBERTa}_{\text{BASE}} \).
}
\end{table}

%% file: main.bbl
\begin{thebibliography}{42}
\expandafter\ifx\csname natexlab\endcsname\relax\def\natexlab#1{#1}\fi

\bibitem[{Belinkov(2022)}]{probing_belinkov}
Yonatan Belinkov. 2022.
\newblock \href {https://doi.org/10.1162/coli_a_00422} {{Probing Classifiers:
  Promises, Shortcomings, and Advances}}.
\newblock \emph{Computational Linguistics}, 48(1):207--219.

\bibitem[{Bibal et~al.(2022)Bibal, Cardon, Alfter, Wilkens, Wang,
  Fran{\c{c}}ois, and Watrin}]{bibal-etal-2022-attention}
Adrien Bibal, R{\'e}mi Cardon, David Alfter, Rodrigo Wilkens, Xiaoou Wang,
  Thomas Fran{\c{c}}ois, and Patrick Watrin. 2022.
\newblock \href {https://doi.org/10.18653/v1/2022.acl-long.269} {Is attention
  explanation? an introduction to the debate}.
\newblock In \emph{Proceedings of the 60th Annual Meeting of the Association
  for Computational Linguistics (Volume 1: Long Papers)}, pages 3889--3900,
  Dublin, Ireland. Association for Computational Linguistics.

\bibitem[{Brown et~al.(2020)Brown, Mann, Ryder, Subbiah, Kaplan, Dhariwal,
  Neelakantan, Shyam, Sastry, Askell et~al.}]{gpt3}
Tom Brown, Benjamin Mann, Nick Ryder, Melanie Subbiah, Jared~D Kaplan, Prafulla
  Dhariwal, Arvind Neelakantan, Pranav Shyam, Girish Sastry, Amanda Askell,
  et~al. 2020.
\newblock \href
  {https://proceedings.neurips.cc/paper/2020/hash/1457c0d6bfcb4967418bfb8ac142f64a-Abstract.html}
  {Language models are few-shot learners}.
\newblock \emph{Advances in neural information processing systems},
  33:1877--1901.

\bibitem[{Choromanski et~al.(2021)Choromanski, Likhosherstov, Dohan, Song,
  Gane, Sarl{\'{o}}s, Hawkins, Davis, Mohiuddin, Kaiser, Belanger, Colwell, and
  Weller}]{performer}
Krzysztof Choromanski, Valerii Likhosherstov, David Dohan, Xingyou Song,
  Andreea Gane, Tam{\'{a}}s Sarl{\'{o}}s, Peter Hawkins, Jared Davis, Afroz
  Mohiuddin, Lukasz Kaiser, David Belanger, Lucy Colwell, and Adrian Weller.
  2021.
\newblock \href {https://arxiv.org/abs/2009.14794} {Rethinking attention with
  {Performers}}.
\newblock In \emph{Proc.\ of ICLR}.

\bibitem[{Chowdhery et~al.(2022)Chowdhery, Narang, Devlin, Bosma, Mishra,
  Roberts, Barham, Chung, Sutton, Gehrmann, Schuh, Shi, Tsvyashchenko, Maynez,
  Rao, Barnes, Tay, Shazeer, Prabhakaran, Reif, Du, Hutchinson, Pope, Bradbury,
  Austin, Isard, Gur-Ari, Yin, Duke, Levskaya, Ghemawat, Dev, Michalewski,
  Garcia, Misra, Robinson, Fedus, Zhou, Ippolito, Luan, Lim, Zoph, Spiridonov,
  Sepassi, Dohan, Agrawal, Omernick, Dai, Pillai, Pellat, Lewkowycz, Moreira,
  Child, Polozov, Lee, Zhou, Wang, Saeta, Diaz, Firat, Catasta, Wei,
  Meier-Hellstern, Eck, Dean, Petrov, and Fiedel}]{palm}
Aakanksha Chowdhery, Sharan Narang, Jacob Devlin, Maarten Bosma, Gaurav Mishra,
  Adam Roberts, Paul Barham, Hyung~Won Chung, Charles Sutton, Sebastian
  Gehrmann, Parker Schuh, Kensen Shi, Sasha Tsvyashchenko, Joshua Maynez,
  Abhishek Rao, Parker Barnes, Yi~Tay, Noam Shazeer, Vinodkumar Prabhakaran,
  Emily Reif, Nan Du, Ben Hutchinson, Reiner Pope, James Bradbury, Jacob
  Austin, Michael Isard, Guy Gur-Ari, Pengcheng Yin, Toju Duke, Anselm
  Levskaya, Sanjay Ghemawat, Sunipa Dev, Henryk Michalewski, Xavier Garcia,
  Vedant Misra, Kevin Robinson, Liam Fedus, Denny Zhou, Daphne Ippolito, David
  Luan, Hyeontaek Lim, Barret Zoph, Alexander Spiridonov, Ryan Sepassi, David
  Dohan, Shivani Agrawal, Mark Omernick, Andrew~M. Dai,
  Thanumalayan~Sankaranarayana Pillai, Marie Pellat, Aitor Lewkowycz, Erica
  Moreira, Rewon Child, Oleksandr Polozov, Katherine Lee, Zongwei Zhou, Xuezhi
  Wang, Brennan Saeta, Mark Diaz, Orhan Firat, Michele Catasta, Jason Wei,
  Kathy Meier-Hellstern, Douglas Eck, Jeff Dean, Slav Petrov, and Noah Fiedel.
  2022.
\newblock \href {https://doi.org/10.48550/ARXIV.2204.02311} {{PaLM}: Scaling
  language modeling with pathways}.
\newblock {arXiv}:2204.02311.

\bibitem[{Clark et~al.(2019)Clark, Khandelwal, Levy, and
  Manning}]{clark-etal-2019-bert}
Kevin Clark, Urvashi Khandelwal, Omer Levy, and Christopher~D. Manning. 2019.
\newblock \href {https://aclanthology.org/W19-4828/} {What does {BERT} look at?
  an analysis of {BERT}{'}s attention}.
\newblock In \emph{Proc.\ of BlackboxNLP}.

\bibitem[{Clark et~al.(2020)Clark, Luong, Le, and Manning}]{clark2020electra}
Kevin Clark, Minh-Thang Luong, Quoc~V. Le, and Christopher~D. Manning. 2020.
\newblock \href {https://openreview.net/pdf?id=r1xMH1BtvB} {{ELECTRA}:
  Pre-training text encoders as discriminators rather than generators}.
\newblock In \emph{Proc.\ of ICLR}.

\bibitem[{Devlin et~al.(2019)Devlin, Chang, Lee, and
  Toutanova}]{devlins2019bert}
Jacob Devlin, Ming-Wei Chang, Kenton Lee, and Kristina Toutanova. 2019.
\newblock \href {https://arxiv.org/abs/810.04805} {{BERT}: Pre-training of deep
  bidirectional transformers for language understanding}.
\newblock In \emph{Proc. of NAACL}.

\bibitem[{Dolan and Brockett(2005)}]{dolan-brockett-2005-automatically}
William~B. Dolan and Chris Brockett. 2005.
\newblock \href {https://aclanthology.org/I05-5002} {Automatically constructing
  a corpus of sentential paraphrases}.
\newblock In \emph{Proc.\ of IWP}.

\bibitem[{He et~al.(2021)He, Liu, Gao, and Chen}]{deberta}
Pengcheng He, Xiaodong Liu, Jianfeng Gao, and Weizhu Chen. 2021.
\newblock \href {https://openreview.net/forum?id=XPZIaotutsD} {{DeBERTa}:
  decoding-enhanced bert with disentangled attention}.
\newblock In \emph{Proc.\ of ICLR}.

\bibitem[{Htut et~al.(2019)Htut, Phang, Bordia, and Bowman}]{Htut2019DoAH}
Phu~Mon Htut, Jason Phang, Shikha Bordia, and Samuel~R. Bowman. 2019.
\newblock \href {https://doi.org/10.48550/ARXIV.1911.12246} {Do attention heads
  in bert track syntactic dependencies?}
\newblock {arXiv}:1911.12246.

\bibitem[{Hua et~al.(2022)Hua, Dai, Liu, and Le}]{flash}
Weizhe Hua, Zihang Dai, Hanxiao Liu, and Quoc~V. Le. 2022.
\newblock \href {https://doi.org/10.48550/ARXIV.2202.10447} {Transformer
  quality in linear time}.
\newblock {arXiv}:2202.10447.

\bibitem[{Jain and Wallace(2019)}]{jain-wallace-2019-attention}
Sarthak Jain and Byron~C. Wallace. 2019.
\newblock \href {https://doi.org/10.18653/v1/N19-1357} {{A}ttention is not
  {E}xplanation}.
\newblock In \emph{Proceedings of the 2019 Conference of the North {A}merican
  Chapter of the Association for Computational Linguistics: Human Language
  Technologies, Volume 1 (Long and Short Papers)}, pages 3543--3556,
  Minneapolis, Minnesota. Association for Computational Linguistics.

\bibitem[{Kasai et~al.(2021)Kasai, Peng, Zhang, Yogatama, Ilharco, Pappas, Mao,
  Chen, and Smith}]{kasai2021t2r}
Jungo Kasai, Hao Peng, Yizhe Zhang, Dani Yogatama, Gabriel Ilharco, Nikolaos
  Pappas, Yi~Mao, Weizhu Chen, and Noah~A. Smith. 2021.
\newblock \href {https://arxiv.org/abs/2103.13076} {Finetuning pretrained
  transformers into {RNN}s}.
\newblock In \emph{Proc. of EMNLP}.

\bibitem[{Kitaev et~al.(2020)Kitaev, Kaiser, and
  Levskaya}]{Kitaev2020ReformerTE}
Nikita Kitaev, Lukasz Kaiser, and Anselm Levskaya. 2020.
\newblock \href {https://arxiv.org/abs/2001.04451} {Reformer: The efficient
  transformer}.
\newblock In \emph{Proc.\ of ICLR}.

\bibitem[{Lee-Thorp et~al.(2021)Lee-Thorp, Ainslie, Eckstein, and
  Ontanon}]{fnet}
James Lee-Thorp, Joshua Ainslie, Ilya Eckstein, and Santiago Ontanon. 2021.
\newblock \href {https://doi.org/10.48550/ARXIV.2105.03824} {Fnet: Mixing
  tokens with fourier transforms}.
\newblock {arXiv}:2105.03824.

\bibitem[{Li et~al.(2021)Li, Cotterell, and Sachan}]{Li2021DifferentiableSP}
Jiaoda Li, Ryan Cotterell, and Mrinmaya Sachan. 2021.
\newblock Differentiable subset pruning of transformer heads.
\newblock \emph{Transactions of the Association for Computational Linguistics},
  9:1442--1459.

\bibitem[{Liu et~al.(2021)Liu, Dai, So, and Le}]{gmlp}
Hanxiao Liu, Zihang Dai, David So, and Quoc~V Le. 2021.
\newblock \href
  {https://proceedings.neurips.cc/paper/2021/file/4cc05b35c2f937c5bd9e7d41d3686fff-Paper.pdf}
  {Pay attention to {MLP}s}.
\newblock In \emph{Advances in Neural Information Processing Systems},
  volume~34, pages 9204--9215. Curran Associates, Inc.

\bibitem[{Liu et~al.(2019{\natexlab{a}})Liu, Gardner, Belinkov, Peters, and
  Smith}]{probing_nelson}
Nelson~F. Liu, Matt Gardner, Yonatan Belinkov, Matthew~E. Peters, and Noah~A.
  Smith. 2019{\natexlab{a}}.
\newblock \href {https://arxiv.org/abs/1903.08855} {Linguistic knowledge and
  transferability of contextual representations}.
\newblock In \emph{Proc.\ of NAACL}.

\bibitem[{Liu et~al.(2019{\natexlab{b}})Liu, Ott, Goyal, Du, Joshi, Chen, Levy,
  Lewis, Zettlemoyer, and Stoyanov}]{Liu2019RoBERTaAR}
Yinhan Liu, Myle Ott, Naman Goyal, Jingfei Du, Mandar Joshi, Danqi Chen, Omer
  Levy, Mike Lewis, Luke~S. Zettlemoyer, and Veselin Stoyanov.
  2019{\natexlab{b}}.
\newblock \href {https://arxiv.org/abs/1907.11692} {{RoBERTa}: A robustly
  optimized {BERT} pretraining approach}.
\newblock {arXiv}:1907.11692.

\bibitem[{Merity et~al.(2017)Merity, Xiong, Bradbury, and Socher}]{wiki103}
Stephen Merity, Caiming Xiong, James Bradbury, and Richard Socher. 2017.
\newblock \href {https://arxiv.org/abs/1609.07843} {Pointer sentinel mixture
  models}.
\newblock In \emph{Proc.\ of ICLR}.

\bibitem[{Michel et~al.(2019)Michel, Levy, and Neubig}]{Michel2019AreSH}
Paul Michel, Omer Levy, and Graham Neubig. 2019.
\newblock Are sixteen heads really better than one?
\newblock In \emph{NeurIPS}.

\bibitem[{Peng et~al.(2022)Peng, Kasai, Pappas, Yogatama, Wu, Kong, Schwartz,
  and Smith}]{peng-etal-2022-abc}
Hao Peng, Jungo Kasai, Nikolaos Pappas, Dani Yogatama, Zhaofeng Wu, Lingpeng
  Kong, Roy Schwartz, and Noah Smith. 2022.
\newblock \href {https://aclanthology.org/2022.acl-long.515} {{ABC}: Attention
  with bounded-memory control}.
\newblock In \emph{Proceedings of the 60th Annual Meeting of the Association
  for Computational Linguistics (Volume 1: Long Papers)}, pages 7469--7483,
  Dublin, Ireland. Association for Computational Linguistics.

\bibitem[{Peng et~al.(2020{\natexlab{a}})Peng, Pappas, Yogatama, Schwartz,
  Smith, and Kong}]{RFA}
Hao Peng, Nikolaos Pappas, Dani Yogatama, Roy Schwartz, Noah Smith, and
  Lingpeng Kong. 2020{\natexlab{a}}.
\newblock \href
  {https://openreview.net/forum?continueFlag=7fcad3444ce73135efa053c0f6709de5&id=QtTKTdVrFBB}
  {Random feature attention}.
\newblock In \emph{International Conference on Learning Representations}.

\bibitem[{Peng et~al.(2020{\natexlab{b}})Peng, Schwartz, Li, and
  Smith}]{peng-etal-2020-mixture}
Hao Peng, Roy Schwartz, Dianqi Li, and Noah~A. Smith. 2020{\natexlab{b}}.
\newblock \href {https://doi.org/10.18653/v1/2020.acl-main.587} {A mixture of h
  - 1 heads is better than h heads}.
\newblock In \emph{Proceedings of the 58th Annual Meeting of the Association
  for Computational Linguistics}, pages 6566--6577, Online. Association for
  Computational Linguistics.

\bibitem[{Qin et~al.(2022)Qin, Sun, Deng, Li, Wei, Lv, Yan, Kong, and
  Zhong}]{cosformer}
Zhen Qin, Weixuan Sun, Hui Deng, Dongxu Li, Yunshen Wei, Baohong Lv, Junjie
  Yan, Lingpeng Kong, and Yiran Zhong. 2022.
\newblock \href {https://doi.org/10.48550/ARXIV.2202.08791} {cosformer:
  Rethinking softmax in attention}.
\newblock {arXiv}:2202.08791.

\bibitem[{Raffel et~al.(2020)Raffel, Shazeer, Roberts, Lee, Narang, Matena,
  Zhou, Li, and Liu}]{2020t5}
Colin Raffel, Noam Shazeer, Adam Roberts, Katherine Lee, Sharan Narang, Michael
  Matena, Yanqi Zhou, Wei Li, and Peter~J. Liu. 2020.
\newblock \href {http://jmlr.org/papers/v21/20-074.html} {Exploring the limits
  of transfer learning with a unified text-to-text transformer}.
\newblock \emph{JMLR}.

\bibitem[{Ravishankar et~al.(2021)Ravishankar, Kulmizev, Abdou, Søgaard, and
  Nivre}]{Ravishankar2021AttentionCR}
Vinit Ravishankar, Artur Kulmizev, Mostafa Abdou, Anders Søgaard, and Joakim
  Nivre. 2021.
\newblock \href {https://doi.org/10.48550/ARXIV.2101.10927} {Attention can
  reflect syntactic structure (if you let it)}.
\newblock {arXiv}:2101.10927.

\bibitem[{Schlag et~al.(2021)Schlag, Irie, and Schmidhuber}]{schlag2021linear}
Imanol Schlag, Kazuki Irie, and Jürgen Schmidhuber. 2021.
\newblock \href {https://arxiv.org/abs/2102.11174} {Linear transformers are
  secretly fast weight memory systems}.
\newblock In \emph{Proc.\ of ICML}.

\bibitem[{Serrano and Smith(2019)}]{serrano-smith-2019-attention}
Sofia Serrano and Noah~A. Smith. 2019.
\newblock \href {https://doi.org/10.18653/v1/P19-1282} {Is attention
  interpretable?}
\newblock In \emph{Proceedings of the 57th Annual Meeting of the Association
  for Computational Linguistics}, pages 2931--2951, Florence, Italy.
  Association for Computational Linguistics.

\bibitem[{Socher et~al.(2013)Socher, Perelygin, Wu, Chuang, Manning, Ng, and
  Potts}]{socher-etal-2013-recursive}
Richard Socher, Alex Perelygin, Jean Wu, Jason Chuang, Christopher~D. Manning,
  Andrew Ng, and Christopher Potts. 2013.
\newblock \href {https://aclanthology.org/D13-1170/} {Recursive deep models for
  semantic compositionality over a sentiment treebank}.
\newblock In \emph{Proc.\ of EMNLP}.

\bibitem[{Tjong Kim~Sang and
  De~Meulder(2003)}]{tjong-kim-sang-de-meulder-2003-introduction}
Erik~F. Tjong Kim~Sang and Fien De~Meulder. 2003.
\newblock \href {https://aclanthology.org/W03-0419} {Introduction to the
  {C}o{NLL}-2003 shared task: Language-independent named entity recognition}.
\newblock In \emph{Proc.\ of CoNLL}.

\bibitem[{Vaswani et~al.(2017)Vaswani, Shazeer, Parmar, Uszkoreit, Jones,
  Gomez, Kaiser, and Polosukhin}]{Vaswani2017AttentionIA}
Ashish Vaswani, Noam Shazeer, Niki Parmar, Jakob Uszkoreit, Llion Jones,
  Aidan~N. Gomez, \L{}ukasz Kaiser, and Illia Polosukhin. 2017.
\newblock \href {https://arxiv.org/pdf/1706.03762.pdf} {Attention is all you
  need}.
\newblock In \emph{Proc. of NeurIPS}.

\bibitem[{Voita et~al.(2019)Voita, Talbot, Moiseev, Sennrich, and
  Titov}]{voita-etal-2019}
Elena Voita, David Talbot, Fedor Moiseev, Rico Sennrich, and Ivan Titov. 2019.
\newblock \href {https://doi.org/10.18653/v1/P19-1580} {Analyzing multi-head
  self-attention: Specialized heads do the heavy lifting, the rest can be
  pruned}.
\newblock In \emph{Proceedings of the 57th Annual Meeting of the Association
  for Computational Linguistics}, pages 5797--5808, Florence, Italy.
  Association for Computational Linguistics.

\bibitem[{Wang et~al.(2019)Wang, Singh, Michael, Hill, Levy, and Bowman}]{GLUE}
Alex Wang, Amanpreet Singh, Julian Michael, Felix Hill, Omer Levy, and
  Samuel~R. Bowman. 2019.
\newblock \href {https://arxiv.org/abs/1804.07461} {{GLUE:} {A} multi-task
  benchmark and analysis platform for natural language understanding}.
\newblock In \emph{Proc.\ of ICLR}.

\bibitem[{Wang et~al.(2022)Wang, Chen, Ren, Liang, Yan, and
  Ren}]{Wang2022paying}
Shanshan Wang, Zhumin Chen, Zhaochun Ren, Huasheng Liang, Qiang Yan, and
  Pengjie Ren. 2022.
\newblock \href {https://doi.org/10.48550/ARXIV.2204.02922} {Paying more
  attention to self-attention: Improving pre-trained language models via
  attention guiding}.
\newblock {arXiv}:2204.02922.

\bibitem[{Wang et~al.(2020)Wang, Li, Khabsa, Fang, and
  Ma}]{Wang2020LinformerSW}
Sinong Wang, Belinda~Z. Li, Madian Khabsa, Han Fang, and Hao Ma. 2020.
\newblock \href {https://doi.org/10.48550/ARXIV.2006.04768} {Linformer:
  Self-attention with linear complexity}.
\newblock {arXiv}:2006.04768.

\bibitem[{Warstadt et~al.(2019)Warstadt, Singh, and
  Bowman}]{warstadt-etal-2019-neural}
Alex Warstadt, Amanpreet Singh, and Samuel~R. Bowman. 2019.
\newblock \href {https://aclanthology.org/Q19-1040} {Neural network
  acceptability judgments}.
\newblock \emph{TACL}.

\bibitem[{Wiegreffe and Pinter(2019)}]{wiegreffe-pinter-2019-attention}
Sarah Wiegreffe and Yuval Pinter. 2019.
\newblock \href {https://doi.org/10.18653/v1/D19-1002} {Attention is not not
  explanation}.
\newblock In \emph{Proceedings of the 2019 Conference on Empirical Methods in
  Natural Language Processing and the 9th International Joint Conference on
  Natural Language Processing (EMNLP-IJCNLP)}, pages 11--20, Hong Kong, China.
  Association for Computational Linguistics.

\bibitem[{Williams et~al.(2018)Williams, Nangia, and
  Bowman}]{williams-etal-2018-broad}
Adina Williams, Nikita Nangia, and Samuel Bowman. 2018.
\newblock \href {https://arxiv.org/abs/1704.05426} {A broad-coverage challenge
  corpus for sentence understanding through inference}.
\newblock In \emph{Proc.\ of NAACL}.

\bibitem[{Wolf et~al.(2020)Wolf, Debut, Sanh, Chaumond, Delangue, Moi, Cistac,
  Rault, Louf, Funtowicz, Davison, Shleifer, von Platen, Ma, Jernite, Plu, Xu,
  Scao, Gugger, Drame, Lhoest, and Rush}]{wolf-etal-2020-transformers}
Thomas Wolf, Lysandre Debut, Victor Sanh, Julien Chaumond, Clement Delangue,
  Anthony Moi, Pierric Cistac, Tim Rault, Rémi Louf, Morgan Funtowicz, Joe
  Davison, Sam Shleifer, Patrick von Platen, Clara Ma, Yacine Jernite, Julien
  Plu, Canwen Xu, Teven~Le Scao, Sylvain Gugger, Mariama Drame, Quentin Lhoest,
  and Alexander~M. Rush. 2020.
\newblock \href {https://www.aclweb.org/anthology/2020.emnlp-demos.6}
  {Transformers: State-of-the-art natural language processing}.
\newblock In \emph{Proceedings of the 2020 Conference on Empirical Methods in
  Natural Language Processing: System Demonstrations}, pages 38--45, Online.
  Association for Computational Linguistics.

\bibitem[{You et~al.(2020)You, Sun, and Iyyer}]{gauss_mt}
Weiqiu You, Simeng Sun, and Mohit Iyyer. 2020.
\newblock \href {https://doi.org/10.48550/ARXIV.2005.00742} {Hard-coded
  gaussian attention for neural machine translation}.
\newblock {arXiv}:2005.00742.

\end{thebibliography}
